# Phonetic Transfer of /i/ from Mandarin Chinese to General American English: A Case Study of a Chinese Learner with Advanced English

Lintao Chen[1,*]

[1]Teachers College, Columbia University, New York, the United States, NY 10027
[*]Corresponding author. Email: lc3590@tc.columbia.edu

**ABSTRACT**

The current paper concerns language transfer at the phonetic level and concentrates on the transfer phenomenon in an advanced English language learner's acquisition of the English vowels /i/ and /ɪ/. By determining whether the Chinese English-language learner (ELL), named Vanya, can accurately distinguish between /i/ and /ɪ/, and pronounce them precisely in General American English (GAE), this paper serves as a reference for further studying language transfer among Chinese ELLs. There were two objectives: first, the learner's perceptual ability to distinguish between vowels /i/ and /ɪ/ was examined; second, the effect of the phonetic transfer was determined. Two perception tests and a production test were used to attain these two objectives. The results of two perception tests demonstrated Vanya's perceptual competence in distinguishing between /i/ and /ɪ/ and laid a solid foundation for the validity of the subsequent production test. Given that Vanya's production of F1 and F2 values of /i/ were highly similar across his first language (Mandarin Chinese) and second language (GAE) and that both values were lower than the typical values for common /i/ in GAE, with an especially prominent disparity between the F2 values, it is reasonable to conclude that a phonetic transfer occurred. The participant's high perceptual competence as an advanced-level ELL did not noticeably moderate the effect of phonetic transfer.

## 1. INTRODUCTION

Linguists, including Charles Fries, Edward Sapir, Charles Hockett, and Burrhus Skinner, conducted research in the 1950s that led to the development of the contrastive analysis (CA) technique, which focuses on comparing and contrasting two language systems of second language learners [1]. According to Richards and Schmidt (2010), CA is built on a number of assumptions, one of which is that the CA technique can predict which challenges a group of language learners might have [2]. For example, Chinese English-language learners (ELLs) may struggle to pronounce the English consonant /v/ since the sound does not exist in Chinese. In many cases, the /v/ sound is replaced with /w/.

Such mispronunciations occur due to what is known as negative transfer. Negative transfer - defined as "an improper influence of the first language structure or regulation on the second language use" - is one of CA's basic assumptions [3]. Since this realization, the phenomenon of language transfer has been the focus of myriad researchers, and even now, when CA has long been overshadowed by new theories, studies on language transfer have been continuously carried out. It is generally acknowledged that the first language (L1) does affect the acquisition of the second language (L2).

This paper concerns language transfer at the phonetic level and focuses on the transfer phenomenon existing in advanced-level Chinese ELL acquisition of the English vowels /i/ and /ɪ/. In the Chinese pronunciation system, the tense high front vowel /i/ does exist; however, its lax counterpart /ɪ/ does not. Under this circumstance, a phonetic transfer might happen.

While there are studies addressing the issues of Chinese ELLs making a phonetic transfer from Chinese to English vowel systems, very few of them have studied the phonetic transfer of vowels among advanced ELLs. In this case, by determining whether an advanced Chinese ELL, named Vanya, can accurately distinguish between /i/ and /ɪ/ and pronounce them in English words precisely, this paper serves as a reference for further studying the language transfer of Chinese ELLs [4].

## 2. PHONETIC TRANSFER OF /I/ FROM MANDARIN CHINESE TO GENERAL AMERICAN ENGLISH

### 2.1. Background

This study focuses on the general American English /i/ and /ɪ/ because of the increased exposure of Chinese students to American English. For instance, in the Chinese College English Test (the national Chinese test for English as a foreign language, with Band 6 as the highest level) in Band 4 and Band 6, listening materials in American English prevail. The situation not only suggests that the status of American English is widely recognized in China but also encourages Chinese ELLs to take American English as a subject of study.

The current paper examines this learning process at the phonetic level and focuses on the acquisition of specific vowels. It is foreseeable that the number of American English learners will continue to boom in the future, and research on Chinese learners' second language acquisition process of American English will continue to be conducted.

### 2.2. Mandarin Chinese /i/

There is no definitive conclusion about the number of Mandarin Chinese vowels. This paper adopts Duanmu's (2007) argument, which claims that Chinese possesses 13 vowels ([i], [y], [u], [o], [E], [ɤ], [e], [ə], [A], [ɑ], [a], [æ], [ɐ]) formed from 5 vowel phonemes (/i/, /y/, /u/, /ə/, /a/), with the exception of the two "apical vowels" and the retroflex vowel [ɚ]) [4]. For the purpose of this study, /i/ (as in *"必"* /bi/) is characterized as the high front tense vowel.

### 2.3. General American English /i/ and /ɪ/

American English pronunciation can vary from region to region in the United States. The American English studied in this paper is based on the criteria proposed by Kenyon and Knott [5]. There are 17 vowels in American English: [i], [ɪ], [e], [ɛ], [æ], [a], [ɝ], [ɜ], [ɚ], [ə], [ɑ], [ɒ], [ʌ], [ɔ], [o], [ʊ] and [u]. Among these vowels, /i/ (as in "*beat*") is the high front tense vowel, and /ɪ/ (as in "*bit*") is the high front lax vowel. These two vowels share a certain degree of similarity; therefore, mastering them might be challenging for ELLs.

### 2.4. Language Transfer

According to Richards and Schmidt (2010), CA is built on the widely accepted language transfer theory, which has two forms: positive transfer and negative transfer. Positive transfer is defined by Dulay, Burt, and Krashen as "the automatic use of the L1 structure in L2 performance when both languages' structures are the same, resulting in correct utterances". Positive transfer occurs when the native language and the target language share some linguistic components in common. Negative transfer, also known as interference, occurs as a result of the discrepancies between the two language systems. Such disparities make learning the target language difficult and increase the likelihood of the learner making mistakes, producing an interlanguage [6].

However, CA was challenged by the notion that language acquisition is an active rule-building behavior rather than a habit-forming behavior, which eventually led to its demise [7]. Nevertheless, research on language transfer has not ceased. In various new theoretical frameworks and models, language transfer possesses the same important status as in CA. For instance, in error analysis, linguistic interference, according to Corder (1973), is the clearest evidence to explain learners' errors [8]. Many other studies also rely on the premise that learners' errors can be caused by transferring L1 elements into the L2.

The present study represents an attempt to focus on the language transfer at the phonetic level of advanced ELLs with Chinese as the L1. Through the analysis of the results from two perception tests and one production test, this article aims to provide a useful reference for future studies of the language transfer in Chinese ELLs.

## 3. METHODOLOGY

The phonetic transfer of the vowel /i/ from Mandarin Chinese to American English vowels /i/ and /ɪ/ is the subject of this thesis. Due to negative transfer, Chinese ELLs might have more difficulties learning the GAE /i/ and /ɪ/, even if they are advanced learners. To further address the issue, two objectives must be met: first, learners' perceptual ability to distinguish between vowels /i/ and /ɪ/ must be examined; and second, the effect of the phonetic transfer must be determined. Two perception tests and a production test were used to attain these objectives. All the tests were completed by an advanced ELL who had Chinese as an L1.

### 3.1. Subject

The subject, named Vanya, was an adult Chinese learner of English. At the time of this study, he was a graduate student. Vanya was a fluent speaker of GAE and had passed the College English Test Band 6. Throughout Vanya's middle and high school years, he had received a seven-year, nonsubtractive bilingual education, according to the definition given by Ofelia García regarding models of bilingualism [9]. His English learning experience was more comprehensive and in-depth than the overall English acquisition experience in China, which might place emphasis on mastery of English at the written level at the expense of speaking skills.

Vanya speaks Mandarin Chinese as his mother tongue and learns English as a second language. He self-reported that he did not have hearing, reading, or speaking impairments or any other disabilities that might affect his performance on the perception tests and the production test.

### 3.2. Experimental tests

#### 3.2.1. Perception Test 1

This perception test was designed to examine Vanya's ability to perceive the difference between /i/ and /ɪ/ in words. It was a fundamental task for this study to ensure that for the subsequent perception task 2, he already had the basic ability to distinguish between /i/ and /ɪ/ on the perceptual level.

#### 3.2.1.1. Procedures

The material in Table 1 was sent electronically to the participant Vanya. He was presented with one-syllable Chinese and English words containing /i/ or /ɪ/. During the experiment, Vanya typed the vowel of each one-syllable Chinese or English word. His answers were compared with the phonemic transcriptions of the vowels in these words in the Collins dictionary. The participant's ability to identify /i/ and /ɪ/ was determined by the accuracy rate.

**Table 1.** Perception task 1 word set

| mead | did | deed | mid | bid | bead | neat |
|---|---|---|---|---|---|---|
|  |  |  |  |  |  |  |
| 地 (/di/) | 避 (/bi/) | 逆 (/ni/) | Peat | eat | Tim | pit |
|  |  |  |  |  |  |  |
| nit | it | team | 蜜 (/mi/) | 辟 (/pi/) | 替 (/ti/) | 易 (/i/) |
|  |  |  |  |  |  |  |

*Note: The phonetic symbols after the Chinese characters are only used as a reference; the participant was not provided with phonetic symbols at the time of testing.

#### 3.2.2. Perception Test 2

This task was designed to determine the exact degree of similarity Vanya perceives between Chinese /i/ and GAE /i/ and /ɪ/ [10]. As shown in Table 2, for each column, Vanya was expected to compare three one-syllable English words containing /i/ or /ɪ/ with a one-syllable Chinese word containing /i/. To capture the true picture of his interlanguage system as closely as possible, real words and characters were used instead of pure vowels. To optimize the test results, the distribution of words was randomized.

**Table 2.** Perception task 2 word set

| Column 1 | Column 2 | Column 3 | Column 4 |
|---|---|---|---|
| 匿 (/ni/) | 帝 (/di/) | 必 (/bi/) | 剔 (/ti/) |
| meet | nit | myth | peat |
| bid | pit | bead | eat |
| rid | it | neat | beam |

*Note: The phonetic symbols after the Chinese characters are only used as a reference; the participant was not provided with phonetic symbols at the time of testing.

#### 3.2.2.1. Procedures

The material in Table 2 was presented electronically to Vanya. During the experiment, he was asked to listen to English words while comparing each of them with the Chinese character in the same column. Every English word was produced twice. He was then asked to assign a number to each English word according to the similarity of the vowel in that word with that in the reference Chinese character on a scale from zero to five with 5 representing the highest similarity.

All the words Vanya listened to were produced by the General American English recordings from the Collins Online Dictionary.

#### 3.2.3. The Production Test

The production test was designed to examine the effect of phonetic transfer from Mandarin /i/ to American English /i/ and /ɪ/ by acoustic measurements. By comparing Vanya's experimental frequencies as captured by Praat, a software package for sound analysis, with the common varieties of GAE male vowel frequencies [11], a formant-frequency analysis was undertaken to provide direct evidence about language transfer.

#### 3.2.3.1. Procedures

The material in Table 3 was presented electronically to Vanya. During this test, he was asked to pronounce Chinese and English words in his natural way of speaking. His pronunciation of all the words in the table was recorded. Then, the first formant (F1) and the second formant (F2) of the vowels /i/ and /ɪ/ in the words were extracted from his speech with Praat to perform a formant-frequency analysis. Finally, a vowel formant diagram containing the frequencies of Vanya's vowel pronunciation and the common varieties of GAE male high front vowel pronunciation was generated. The effect of phonetic transfer was inferred from the diagram.

**Table 3.** Production task word set

| 辟 (/pi/) | 替 (/ti/) | 易 (/i/) | 地 (/di/) | 避 (/bi/) | 逆 (/ni/) | 蜜 (/mi/) |
|---|---|---|---|---|---|---|
| dip | be | bid | nit | mid | deep | neat |
| me | peak | tea | eat | pick | tip | it |

*Note: The phonetic symbols after the Chinese characters are only used as a reference; the participant was not provided with phonetic symbols at the time of testing.

## 4. RESULTS AND DISCUSSION

### 4.1. Result of perception test 1

Based on a comparison with the phonemic transcription in the Collins dictionary, Vanya's accuracy rate is 100%. This result indicates that he could consciously distinguish between /i/ and /ɪ/, laying the foundation for studying his perceptual ability in the subsequent perception test 2.

### 4.2. Result of perception test 2

Table 4 represents the grades of similarity Vanya perceived between Chinese /i/ and GAE /i/ and /ɪ/. The median value between 0 and 5 is used as the cutoff for similarity. Values greater than or equal to 2.6 are deemed to represent when Vanya considers the two vowels "similar", while other values are taken to mean Vanya perceives the two vowels as "different".

**Table 4.** Perception test 2 result

|  |  | Column 1 | Column 2 | Column 3 | Column 4 | Average Similarity | Similar/Different |
|---|---|---|---|---|---|---|---|
| Chinese /i/ | English /i/ | 5.0 | / | 4.5 | 2.7 | 4.07 | Similar |
| Chinese /i/ | English /ɪ/ | 0 | 3.0 | 2.0 | / | 1.67 | Different |

### 4.3. Results of the production task

#### 4.3.1. Formant-frequency diagram

The result from the formant-frequency analysis for Vanya is shown in Figure 1. His pronunciations of /i/ in Mandarin (Average F1=333.6, Average F2=1592.3) and English (Average F1=314.6, Average F2= 1493.4) words, /ɪ/ in English words (Average F1=456.3, Average F2= 1407.3) as well as the GAE values of /i/ (F1=342, F2=2322) and /ɪ/ (F1=427, F2=2034) for men are plotted.

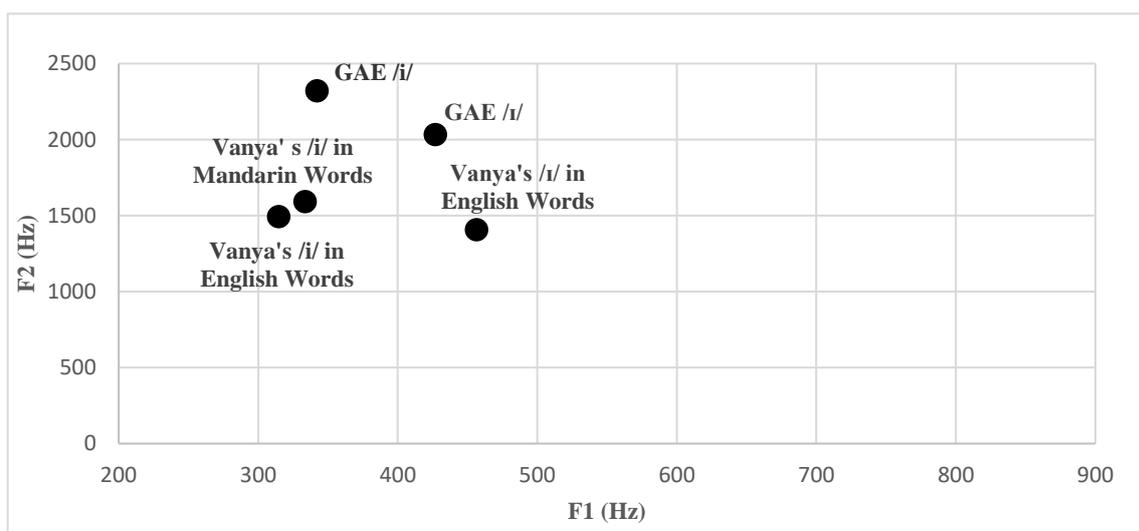

**Figure 1.** Formant-frequency diagram of Vanya's pronunciations

## 5. CONCLUSION

### 5.1. Findings

This study, based on the framework of language transfer theory, examines the phonetic transfer of /i/ from Mandarin Chinese to GAE using the example of an advanced ELL with an L1 of Mandarin Chinese. To address the issue, two objectives were set: first, learners' perceptual ability to distinguish between /i/ and /ɪ/ were examined; and second, the effect of the phonetic transfer was determined. Two perception tests and a production test were designed to attain these two goals.

In perception test 1, participant Vanya was asked to type the vowel he perceived for each one-syllable Chinese or English word provided. By comparison with the phonemic transcription in the Collins dictionary, the accuracy rate (100%) was generated.

In perception test 2, Vanya was asked to listen to English words while comparing them with given Chinese characters. He was then asked to rate the vowel in each English word according to the similarity with the vowel in the reference Chinese character. It was found that Vanya's perception of the similarities is in line with current common consensus on pronunciation. These results affirmed his ability to distinguish between /i/ and /ɪ/ on the perceptual level, and they served as the prerequisite for the validity of the production test's result.

In the production test, given Vanya's competence in recognizing the similarity between Mandarin /i/ and GAE /i/, as well as the fact that both F1 and F2 values of /i/ in his L1 and L2 productions are lower than that of GAE /i/, it was reasonable to conclude that a phonetic transfer occurred. As the disparity among F2 values is prominent, it can be inferred that the transfer from L1 affected Vanya's pronunciation of the high front tense vowel in L2, which led to a backward shift of his tongue position during pronunciation compared to the tongue positions of native GAE speakers. The participant's high perceptual competence as an advanced ELL did not noticeably moderate the effect of phonetic transfer.

Potential evidence supporting that the subject's pronunciation of /ɪ/ in the L2 is possibly influenced by that of /i/ in the L1 can be found in the analysis of the F2 values. Vanya's articulation of Mandarin /i/ and English /ɪ/ is distinctly lower than the GAE /ɪ/ in F2. Nevertheless, this impact of the test taker's phonetic transfer from L1 to L2 in relation to the pronunciation of /ɪ/ cannot be fully determined. Compared with the F1 and F2 values of GAE /ɪ/, the lower F2 values in his production of /ɪ/ is not in accordance with the higher F1 values, considering the F1 and F2 values of his Mandarin /i/ production..

### 5.2. Limitations

Although this study keeps improving its experimental design to make the findings more accurate and reliable, it is undeniable that there are still some limitations.

First, there is no solid definition for an "advanced-level" ELL. The criteria for high proficiency in English in this paper are based on both quantitative indicators, i.e., whether the subject passed College English Test Band 6, and qualitative measurement, i.e., whether the subject was educated in a systematic bilingual manner. Even so, more in-depth assessments are needed to precisely determine the level of language proficiency of the subject.

Second, the sample size of this paper is insufficient. Although a case study ensures the accuracy of the measurement for a specific subject, it downgrades the prominence of the results. Experiments on a larger scale can test the findings.

Finally, a more comprehensive acoustic analysis is needed. The acoustic measurements in this paper focus on the F1 and F2 values of the vowels and do not cover other features. A more thorough acoustic analysis can enhance the significance of the test results.

## ACKNOWLEDGMENTS

I would like to thank Vanya, who agreed to participate in my experiments, and the teachers who supported and encouraged me. Your presence has undoubtedly added to the value of this paper.